\documentclass{article}
\usepackage{csquotes}
\usepackage{spconf,amsmath,graphicx,siunitx,microtype,amsfonts}
\usepackage[numbers]{natbib}
\usepackage{array}
\usepackage{booktabs}
\usepackage{microtype}
\usepackage{url}
\usepackage{amsmath}
\usepackage{siunitx}
\usepackage{color}
\usepackage{multirow}
\usepackage{booktabs}
\usepackage{float}
\usepackage{bbding}

\ninept

\title{Dual Path Modeling for Semantic Matching by Perceiving Subtle Conflicts}
\name{Chao Xue$^{\spadesuit}$\textsuperscript{,1}, Di Liang$^{\clubsuit}$\textsuperscript{,1}, Sirui Wang$^{\clubsuit}$, Jing Zhang$^{\spadesuit}$\textsuperscript{,2}, Wei Wu$^{\clubsuit}$
}
\address{
$^\spadesuit$ School of Software, Beihang University, Beijing, China \\
			$^\clubsuit$Centre for Natural Language Processing, Meituan Inc., Beijing, China
   \\\{xuechao, zhang\_jing\}@buaa.edu.cn , \{liangdi04, wangsirui, wuwei30\}@meituan.com
			}

\begin{document}
% \ninept
%

\maketitle
\footnotetext[1]{Equal contribution.} 
\footnotetext[2]{Corresponding author.} 
\begin{abstract}
Transformer-based pre-trained models have achieved great improvements in semantic matching. However, existing models still suffer from insufficient ability to capture subtle differences.
%Minor noise like word addition, deletion, and modification of sentences may cause flipped predictions. 
The modification, addition and deletion of words in sentence pairs may make it difficult for the model to predict their relationship.
To alleviate this problem, 
%we propose a novel \textbf{Dual Attention Enhanced BERT (DABERT)} to enhance the ability of BERT to capture fine-grained differences in sentence pairs. 
%We propose a novel dual-path modeling mechanism to enhance the model's perception of subtle differences.
%We propose a novel Dual Path Modeling framework to enhance the model's ability to perceive subtle differences in sentence pairs. 
we propose a novel Dual Path Modeling Framework to enhance the model's ability to perceive subtle differences in sentence pairs by separately modeling affinity and difference semantics.
%The key idea is that combining similarities and differences in sentence pairs can better model their matching relationships.
Based on dual-path modeling framework we design the Dual Path Modeling Network (\textbf{DPM-Net}) to recognize semantic relations. 
%To this end, our model employs two distinctly unique components.
%(1) We introduce two different types of contrastive attention functions, focusing on the similarity and dissimilarity in sentences, respectively, and (2) we introduce a composition and representation layer to fuse the tensor obtained by the two different attention.
And we conduct extensive experiments on 10  well-studied semantic matching and robustness test datasets, and the experimental results show that our proposed method achieves consistent improvements over baselines.

\end{abstract}
\section{Introduction}
\label{sec:intro}
Semantic Sentence Matching (SSM) is a fundamental NLP task. 
It's goal is to compare two sentences and identify their semantic relationship. In paraphrase identification, SSM is used to determine whether two sentences are paraphrase or not \cite{madnani-etal-2012-examining}.
In natural language inference task, SSM is utilized to judge whether a hypothesis sentence can be inferred from a premise sentence \cite{bowman2015large}. 
In the answer sentence selection task, SSM is employed to assess the relevance between query-answer pairs and rank all candidate answers \cite{wang2020multi}. 

Across the rich history of semantic sentence matching research, there have been two main streams of studies for solving this problem. One is to utilize a sentence encoder to convert sentences into low-dimensional vectors , and apply a parameterized function to learn the matching scores between them \cite{wang2020multi}.
Another paradigm adopts attention mechanism to calculate scores between tokens from two sentences, and then the matching scores are aggregated to make a sentence-level decision \cite{chen2016enhanced,tay2017compare}.
In recent years, pre-trained models, such as BERT~\cite{devlin2018bert}, RoBERTa~\cite{liu2019roberta}, have became much more popular and achieved outstanding performance in SSM. 
%And they are shown to be powerful contextual representations for predicting sentence relations, as they capture richer linguistic hierarchies in sentences.
%

%例子图
\begin{figure}
\centering
\includegraphics[width=0.47\textwidth]{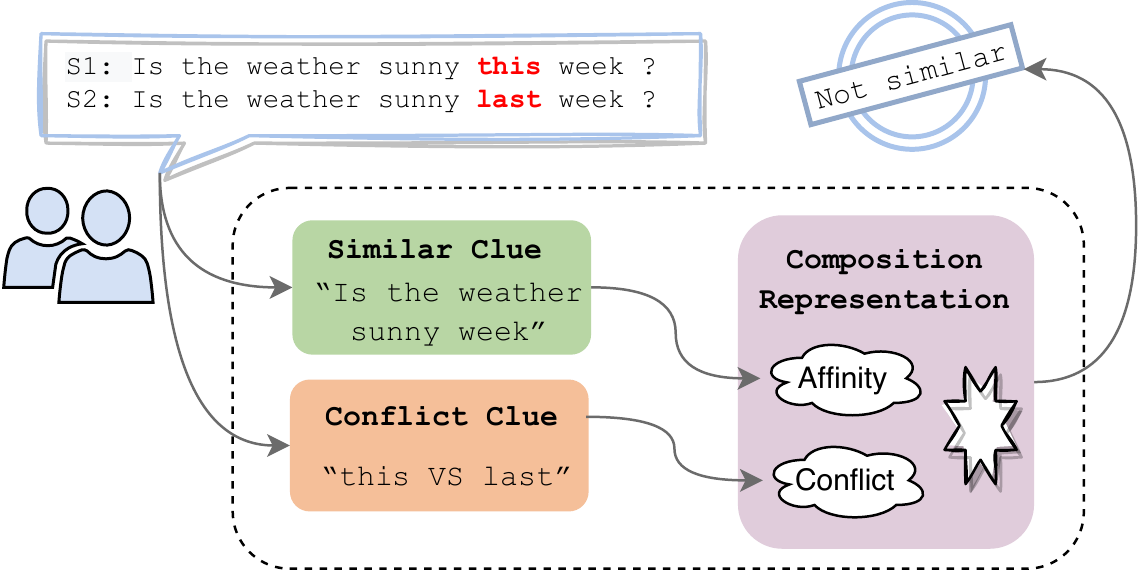}
\caption{\label{fig:example} The Dual Path Modeling Framework for semantic matching.  S1 and S2 are sentence pairs misclassified by BERT.
}
%Example sentences with similar text but different semantics. S1 and S2 are sentence pair. 
\vspace{-0.4cm}
\end{figure}
Although previous studies have provided some insights, existing models still suffer from insufficient ability to capture subtle differences. Figure \ref{fig:example} demonstrates a case suffers from this problem.  Although the sentence pairs in this figure are semantically different, they are too similar in literal for those pre-trained language models to distinguish accurately. 
An important reason is that although the model can measure the matching degree in global semantics, it ignores the local subtle differences between texts. Because for text pairs with highly similar matching words, the overall semantic difference is often caused by different local differences.
Furthermore, existing text matching models based on pretrained models are directly fine-tuned with the training data. It makes the model incapable of generalizing to text matching tasks with highly similar text formats, ultimately resulting in the model lacking the ability to capture fine-grained differences between new samples.
%An important reason is that the two general classification methods are difficult to reflect the subtle differences in sentence pairs: 1)In the representational method: it uses a metric function to evaluate the similarity of two representation vectors, which ignores the semantic difference caused by a few non-core words because most of the words are the same. 2) In the interactive method: it uses \textbf{[CLS]} as the final classification vector. This method focuses on using the context of words to understand the semantics of sentences, without explicitly modeling the semantic differences between sentence pairs, and it is also difficult to reflect subtle differences in sentences. 
Inspired by Sparsegen~\cite{martins2016softmax}, we hypothesize that a more flexible model structure can help the model better understand the relationship of sentence pairs. 
%In this paper, we focus on exploring the common modeling of similarity and dissimilarity of sentence pairs, explicitly modeling dissimilarity and similarity vectors together to further improve the performance of the model. 
%In this paper, we focus on exploring the modeling of inter-text affinities and difference fragments to enhance the model's ability to understand fine-grained semantic differences between texts, thereby improving the performance of text matching tasks. 
In this paper, we focus on exploring the modeling of affinity and difference between texts to enhance the model's ability to understand fine-grained semantic differences, thereby improving the performance of text matching tasks. Therefore, two systemic questions arise naturally:

%\textbf{Q1: How to equip vanilla attention mechanism with the ability on modeling semantics of fine-grained differences between a sentence pair?}
\vspace{0.1cm}
\textbf{Q1: How to equip the model with the ability to model the affinity and difference between sentence pairs?} We analyse that different kinds of attention are complementary clues for sentence matching, which can capture different levels of information in the text sequence. 
%In this paper, we propose a dual attention module including a difference attention accompanied with the affinity attention.
%Affinity attention uses dot-product-based cross-attention to aggregate word- and phrase-level affinity interactions. And the difference attention uses subtraction-based cross-attention to aggregate word- and phrase- level interaction differences. 
%Therefore, we design a dual-path attention mechanism that uses two different attention functions to model the similarity and dissimilarity between texts, and finally obtain semantic representations describing the similarity and dissimilarity, respectively.
In this paper, we propose a dual attention module including a difference attention accompanied with the affinity attention. 
%Affinity attention uses dot-product-based cross-attention to aggregate word- and phrase-level affinity interactions. The difference attention uses subtraction-based cross-attention to aggregate word- and phrase- level interaction differences.and finally obtain semantic representations describing the similarity and dissimilarity, respectively.
Affinity attention and difference attention aggregate word- and phrase-level interactions using dot-product cross-attention and subtraction-based cross-attention. And finally obtain semantic representations describing the affinity and difference, respectively.

%\textbf{Q2: How to design fusion strategies for representations obtained after attention?} 
\vspace{0.1cm}
\textbf{Q2: How to fuse two types of semantic representations into a unified representation?}
We observe that simple aggregation with fixed or average importance weights may be detrimental to fusing heterogeneous vectors. We propose to adaptively aggregate the representations obtained by multiple attention functions from two perspectives. Firstly, the internal aggregation aggregates the matching information together with each word in the sentence in each attention function. Secondly, external aggregation combines the matching information of all attention functions. 
%We apply the aggregation mechanism to adaptively aggregate the two representations.
The output final vectors can better describe the matching details of sentence pairs.

The main contributions of this work can be summarized as follows. First, we conduct an in-depth analysis of the subtle differences in semantic matching and propose a new dual-path modeling framework. Second, the proposed DPM-Net based DPM framework can effectively exploits and aggregates two complementary attention models, such that the intrinsic complex relationship between sentence pairs can be fully discovered for effective semantic matching. Finally, 
%to verify the effectiveness of our proposed model, 
we conduct intensive experiments on 10 matching datasets and robustness testing datasets, and the results show that our method achieves consistent improvements across both architectures (representation-based and interaction-based).

%MRPC QQP SST-B MNLI-m/mm QNLI RTE

\section{Related work}
\label{sec:related_works}
%Our work relates to several work in the literature: semantic Sentence Matching, Robustness test. We will discuss each of these as follows.
\subsection{Semantic Sentence Matching}
Semantic Sentence Matching is a fundamental task in NLP. In recent years, thanks to the appearance of large-scale annotated datasets \cite{bowman2015large}, neural network models have made great progress in SSM \cite{qiu2015convolutional}, mainly fell into two categories. The first one \cite{conneau2017supervised} focuses on encoding sentences into corresponding vector representations without any cross-interaction and applies a classifier layer to obtain similarity. The second one \cite{liang2019asynchronous,chen2016enhanced} utilizes cross-features as an attention module to express the word-level or phrase-level alignments, and aggregates these integrated information to acquire similarity. 
Recently, the shift from neural network architecture engineering to large-scale pre-training has achieved outstanding performance in SSM and many other tasks. Meanwhile, leveraging external knowledge \cite{miller1995wordnet,bodenreider2004unified} to enhance PLMs has been proven to be highly useful for multiple NLP tasks. Therefore, recent work attempts to integrate external knowledge into pre-trained language models, such as AMAN, SemBERT, UERBERT, and so on \cite{liang2019adaptive,zhang2020semantics,liu2023time,xia2021using,bai2021syntax,wang2022dabert,song-etal-2022-improving-semantic}. 
\vspace{-0.1cm}
\subsection{Robustness Test}
Although neural network models have achieved human-like or even superior results in multiple tasks, they still face the insufficient robustness problem in real application scenarios \cite{gui2021textflint}. Tiny literal changes may cause misjudgments. Especially in some cases where fine-grained semantic needs to be discriminated. Besides, most of the current work utilizes one single metric to evaluate their model, may overestimate model capability and lack a fine-grained assessment of model robustness \cite{gui2021textflint}. 
Therefore, recent work starts to focus on robustness research from multiple perspectives. TextFlint \cite{gui2021textflint} incorporates multiple transformations to provide comprehensive robustness analysis. \cite{li2021searching} provide an overall benchmark for current work on adversarial attacks. And \cite{liu2021explainaboard} propose a more comprehensive evaluation system and add more detailed output analysis indicators.

\section{Task Definition}
\label{sec:task definition}
Formally, we can represent each example of sentence pairs as a triple (Q, P, y), where Q = ($q_{1}$,  ..., $q_{N}$) is a sentence with a length N, P = ($p_{1}$, ..., $p_{M}$) is another sentence with a length M, and y $\in$ Y is the label representing the relationship between Q and P. 
%Specifically, for a paraphrase identification task, Q and P are two sentences, Y = 0, 1, where y = 1 means that Q and P are paraphrase of each other, and y = 0 otherwise.
%
Take natural language inference task as an example, Q is a premise sentence, P is a hypothesis sentence,and y=entailment, contradiction, neutral, where entailment indicates P can be inferred from Q, contradiction indicates P cannot be the true condition on Q, and neutral means P and Q are irrelevant to each other. 
%For an answer sentence selection task, Q is a question, P is a candidate answer sentence, and y = {0, 1} where y = 1 means P is a correct answer sentence for Q, and y = 0 otherwise.

%例子图
\begin{figure}[H]
\centering
\includegraphics[width=0.48\textwidth]{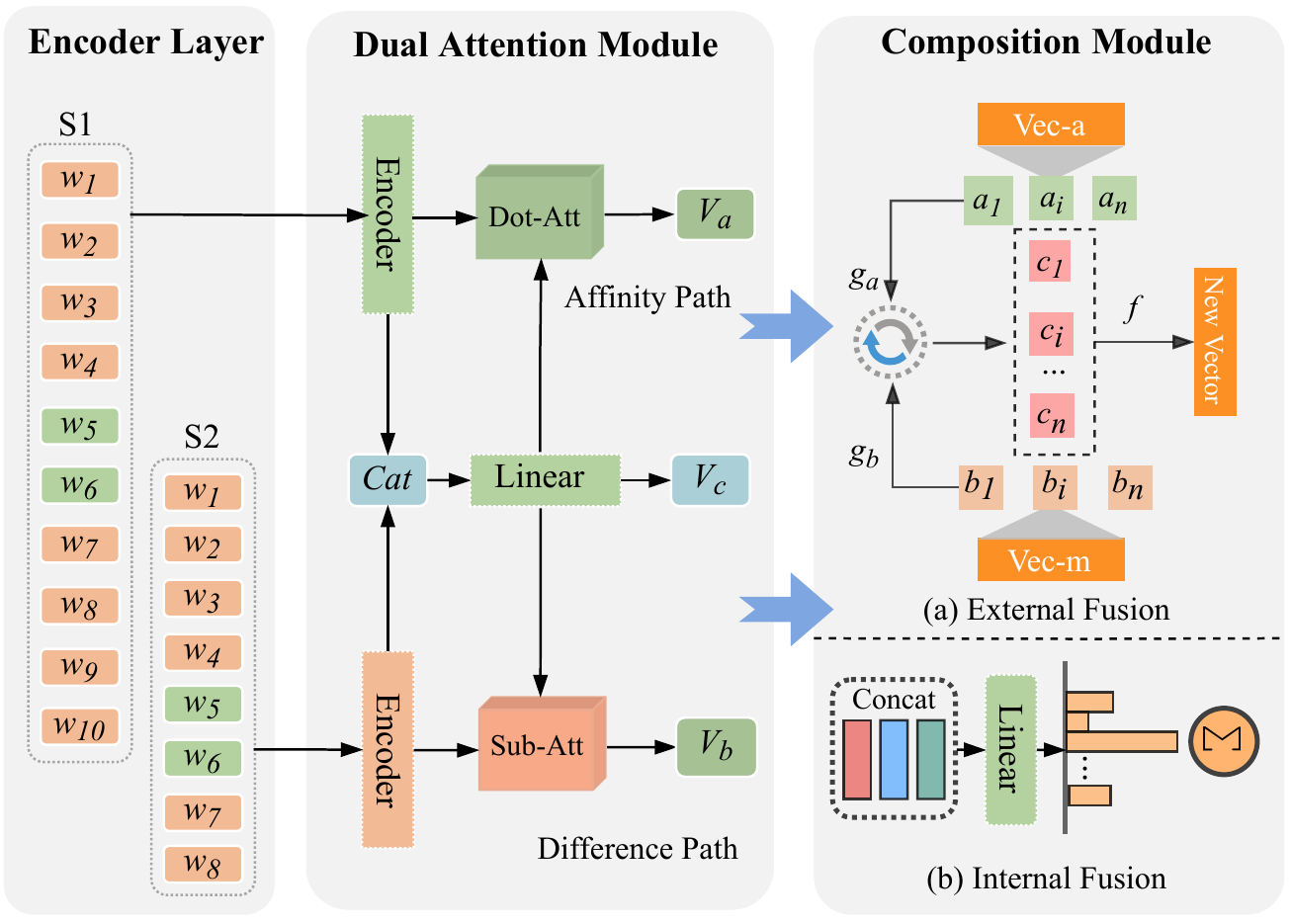}
\caption{\label{fig:module} The overall architecture of the DPM-Net.
}
%Example sentences with similar text but different semantics. S1 and S2 are sentence pair. 
\end{figure}

\section{Method}
\label{sec:method}
We show the design of the Dual Path Modeling Network in Figure \ref{fig:module}. It consists of three parts under the dual path modeling framework. First, We use encoder(eg, transformer, Bert , Roberta) to obtain the context representation of two sentences through representation-based method or interaction-based method. 
Second, we use two different types of attention functions to model the interaction of sentence pairs from different perspectives. Next, we aggregate the matching information along with words in \textbf{P} an \textbf{Q} in two steps. 
We propose to adaptively aggregate representations obtained by dual attention functions from two perspectives. First, internal aggregation aggregates matching information with each word in the sentence in each attention function. Second, external aggregation combines the matching information of all attention features. We apply an aggregation mechanism to adaptively aggregate the two representations.
Finally, we apply a Multilayer Perceptron (MLP) classifier for the final decision.
%get a combined representation vector with more matching details and

%We first match two vectors inside each attention function and then combine the matching information from all functions. The Multi-Layer Perception(MLP) is applied to aggregate the matching information both in the internal aggregation and external aggregation. Finally, we obtain a multi-view attention result after different attention and self-applicable aggregation, Finally, that is, we obtain a multi-view attention result after different attention and self-applicable aggregation, which is used to replace the self-attention part of Bert.

\subsection{Encoder Layer}
For sentence pairs S1=\{${w}_{t}^{p}$\}$_{t=1}^{N}$ and S2=\{${w}_{t}^{q}$\}$_{t=1}^{N}$, we first convert the sentences into vector representations using an encoder. Since we want to explore the performance of DPM on representation-based and interaction-based methods, we use two methods to obtain text representations, respectively. The difference between the two methods is shown in Figure 3, taking Bert as an example. We then use a encoder to produce new representation Q=\{${q}_{1}$,...,${q}_{n}$\} and  P=\{${p}_{1}$,...,${p}_{n}$\} of all words in two sentences respectively.
%\begin{equation}
%\begin{aligned}
%    \label{bert}
%    P &= \textrm{BERT}([\mathrm{CLS}]+S1+[\mathrm{SEP}]) \\
%    Q &= \textrm{BERT}([\mathrm{CLS}]+S2+[\mathrm{SEP}]) \\
%    V &= \textrm{BERT}([\mathrm{CLS}]+S2+[\mathrm{SEP}]) \\
%\end{aligned}
%\end{equation}
%例子图

\begin{figure}[H]
\centering
\includegraphics[width=0.48\textwidth]{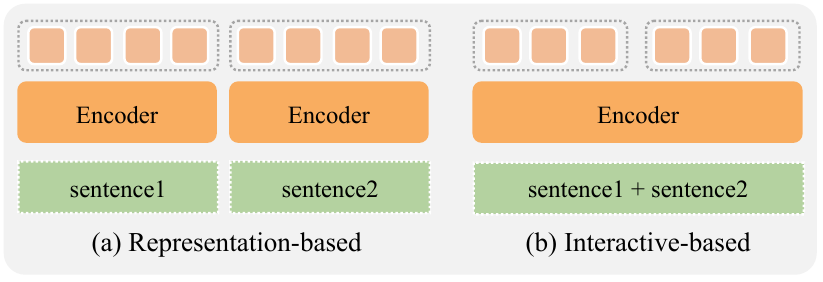}
\caption{\label{fig:encoder} The difference of the two types of encoders.
}
\vspace{-0.2cm}
\end{figure}
At the same time, we concatenate the obtained two representations and perform linear transformation on them, and finally get V=\{${v}_{1}$,...,${v}_{n}$\}, Where N is the length after the text padding.

\begin{table*}
%\vspace{-0.5cm}
\centering
\caption{\label{citation-guide-outsideGlue}
Performance comparison of integrating DPM-Net in \textcolor{red}{interaction-based methods} on 10 Semantic Matching Benchmarks.
%Performance Comparison of 10 Semantic Matching Benchmarks Using DPM-Net on \textcolor{red}{Interactive Method} Based Representation.
%Performance Comparison of 10 Semantic Matching Benchmarks Using DPM-Net in \textcolor{red}{Interactive Methods}
}
\renewcommand\arraystretch{0.9}
\scalebox{0.75}{
\setlength{\tabcolsep}{3.4mm}{
\begin{tabular}{lcccccccccccc}
\toprule
\textbf{Model} & \textbf{Pre-train}&\textbf{MRPC} & \textbf{QQP} & \textbf{STS-B} & \textbf{MNLI-m/mm}& \textbf{QNLI} & \textbf{RTE}  & \textbf{SNLI} & \textbf{Sci} & \textbf{SICK} & \textbf{Twi}  & \textbf{Avg}\\
\hline
%\text{LSTM$\dagger$\cite{hochreiter1997long}$\quad$} & - & - & - & - & - & - & - & - & - & - & -\\
%\textbf{LSTM+DPM(ours)$\dagger$$\quad$}  & \textbf{-} & \textbf{-} & \textbf{-} & \textbf{-} & \textbf{-} & \textbf{-} & \textbf{-} & \textbf{-} & \textbf{-} & \textbf{-}& \textbf{-} \\
%\hline
\text{Transformer$\dagger$\cite{vaswani2017attention}$\quad$} &\XSolidBrush& 81.7 & 84.4 & 70.4 & 72.3/71.4 & 80.3 & 58.1 & 81.7 & 70.6 & - & - & -\\
\textbf{Transformer+DPM(ours)$\dagger$$\quad$}  &\XSolidBrush& \textbf{81.9} & \textbf{85.1} & \textbf{71.8} & \textbf{72.7/72.4} & \textbf{80.9} & \textbf{59.4} & \textbf{85.2} & \textbf{77.3} & \textbf{-} & \textbf{-}& \textbf{-} \\
\hline
\text{BERT-Base$\dagger$\cite{devlin2018bert}$\quad$}&\Checkmark & 87.2 & 89.1 & 87.8 & 84.3/83.7 & 90.4 & 67.2 & 90.7 & 91.8 & 87.2 & 84.8 & 85.8\\
\textbf{BERT-Base+DPM(ours)$\dagger$$\quad$} &\Checkmark & \textbf{89.2} & \textbf{89.5} & \textbf{89.3} & \textbf{85.2/84.8} & \textbf{91.0} & \textbf{68.8} & \textbf{91.2} & \textbf{92.4} & \textbf{87.9} & \textbf{96.6} & \textbf{86.9}\\
\hline
\text{BERT-Large$\dagger$\cite{devlin2018bert}$\quad$}&\Checkmark & 88.9 & 89.3 & 86.6 & 86.8/86.3 & 92.7 & 70.1 & 91.0 & 94.4 & 91.1 & 91.5 & 88.0\\
\textbf{BERT-Large+DPM(ours)$\dagger$$\quad$} &\Checkmark & \textbf{89.6} & \textbf{89.7} & \textbf{88.3} & \textbf{86.9/86.7} & \textbf{93.2} & \textbf{72.5} & \textbf{91.4} & \textbf{94.5} & \textbf{91.4} & \textbf{92.0} & \textbf{88.7}\\
\hline
\text{RoBERTa-Base$\dagger$\cite{liu2019roberta}$\quad$} &\Checkmark& 89.3 & 89.6 & 87.4 & 86.3/86.2 & 92.2 & 73.6 & 90.8 & 92.3 & 87.9 & 85.9 & 87.6 \\
\textbf{RoBERTa-Base+DPM(ours)$\dagger$$\quad$} &\Checkmark & \textbf{89.9} & \textbf{91.0} & \textbf{88.6} & \textbf{87.6/87.2} & \textbf{93.6} & \textbf{81.1} & \textbf{91.5} & \textbf{93.7} & \textbf{89.3} & \textbf{87.3} & \textbf{89.1}\\
\hline
\text{RoBERTa-Large$\dagger$\cite{liu2019roberta}$\quad$} &\Checkmark& 89.4 & 89.7 & 90.2 & 89.5/89.3 & 92.7 & 83.8 & 91.2 & 94.3 & 91.2 & 91.9 & 90.3\\
\textbf{RoBERTa-Large+DPM(ours)$\dagger$$\quad$} &\Checkmark  & \textbf{90.2} & \textbf{91.3} & \textbf{90.8} & \textbf{90.2/90.1} & \textbf{94.0} & \textbf{84.2} & \textbf{91.8} & \textbf{94.8} & \textbf{90.8}  & \textbf{92.3}& \textbf{90.9}\\
\bottomrule
\end{tabular}}}
\vspace{-0.4cm}
\end{table*}

\begin{table*}
%\vspace{-0.5cm}
\centering
\caption{\label{citation-guide-outsideGlue2}
Performance comparison of integrating DPM-Net in \textcolor[rgb]{0.1,0.8,0.9}{representation-based methods} on 10 Semantic Matching Benchmarks.
%Performance Comparison of 10 Semantic Matching Benchmarks Using DPM-Net on \textcolor[rgb]{0.1,0.8,0.9}{Representation Method} Based Representation.
}
\renewcommand\arraystretch{0.9}
\scalebox{0.75}{
\setlength{\tabcolsep}{3.4mm}{
\begin{tabular}{lcccccccccccc}
\toprule
\textbf{Model} & \textbf{Pre-train}&\textbf{MRPC} & \textbf{QQP} & \textbf{STS-B} & \textbf{MNLI-m/mm}& \textbf{QNLI} & \textbf{RTE}  & \textbf{SNLI} & \textbf{Sci} & \textbf{SICK} & \textbf{Twi}  & \textbf{Avg}\\
\hline
%\text{LSTM$\dagger$\cite{hochreiter1997long}$\quad$} & - & - & - & - & - & - & - & - & - & - & -\\
%\textbf{LSTM+DPM(ours)$\dagger$$\quad$}  & \textbf{-} & \textbf{-} & \textbf{-} & \textbf{-} & \textbf{-} & \textbf{-} & \textbf{-} & \textbf{-} & \textbf{-} & \textbf{-}& \textbf{-} \\
%\hline
\text{Transformer$\dagger$\cite{vaswani2017attention}$\quad$} &\XSolidBrush& 71.5 & 79.6 & 66.2 & 66.7/66.5 & 75.8 & 59.2 & 74.3 & 69.9 & - & - & -\\
\textbf{Transformer+DPM(ours)$\dagger$$\quad$}  &\XSolidBrush& \textbf{73.4} & \textbf{83.2} & \textbf{69.4} & \textbf{68.3/68.2} & \textbf{77.7} & \textbf{59.8} & \textbf{80.1} & \textbf{72.4} & \textbf{-} & \textbf{-}& \textbf{-} \\
\hline
\text{BERT-Base$\dagger$\cite{devlin2018bert}$\quad$}&\Checkmark & 81.4 & 82.6 & 81.3 & 78.8/78.4 & 84.4 & 60.2 & 83.3 & 89.8 & 80.9 & 79.1 & 80.4\\
\textbf{BERT-Base+DPM(ours)$\dagger$$\quad$} &\Checkmark & \textbf{83.6} & \textbf{84.4} & \textbf{85.1} & \textbf{79.6/79.4} & \textbf{86.0} & \textbf{63.6} & \textbf{86.1} & \textbf{90.9} & \textbf{82.5} & \textbf{82.7} & \textbf{82.1}\\
\hline
\text{BERT-Large$\dagger$\cite{devlin2018bert}$\quad$}&\Checkmark & 82.5 & 83.4 & 83.8 & 80.3/79.9 & 86.1 & 67.9 & 86.8 & 90.6 & 84.2 & 81.7 & 82.5\\
\textbf{BERT-Large+DPM(ours)$\dagger$$\quad$} &\Checkmark & \textbf{83.3} & \textbf{84.9} & \textbf{86.3} & \textbf{81.6/81.1} & \textbf{87.5} & \textbf{70.1} & \textbf{87.2} & \textbf{91.3} & \textbf{84.9} & \textbf{83.7} & \textbf{83.8}\\
\hline
\text{RoBERTa-Base$\dagger$\cite{liu2019roberta}$\quad$} &\Checkmark& 82.3 & 82.7 & 82.2 & 79.1/78.9 & 85.8 & 65.6 & 84.5 & 90.6 & 82.4 & 81.6 & 81.4 \\
\textbf{RoBERTa-Base+DPM(ours)$\dagger$$\quad$} &\Checkmark & \textbf{83.9} & \textbf{85.2} & \textbf{85.9} & \textbf{79.5/79.3} & \textbf{87.1} & \textbf{67.7} & \textbf{86.8} & \textbf{91.5} & \textbf{82.9} & \textbf{82.8} & \textbf{82.9}\\
\hline
\text{RoBERTa-Large$\dagger$\cite{liu2019roberta}$\quad$} &\Checkmark& 83.4 & 83.8 & 84.2 & 81.3/81.1 & 86.5 & 68.9 & 87.6 & 91.2 & 84.8 & 82.6 & 83.2\\
\textbf{RoBERTa-Large+DPM(ours)$\dagger$$\quad$} &\Checkmark  & \textbf{84.1} & \textbf{85.9} & \textbf{87.5} & \textbf{82.4/82.2} & \textbf{88.0} & \textbf{71.3} & \textbf{88.4} & \textbf{92.5} & \textbf{85.6}  & \textbf{84.8}& \textbf{84.7}\\
\bottomrule 
\end{tabular}}}
\vspace{-0.4cm}
\end{table*}

\subsection{Dual Attention Module}
In dual attention module, we use two different attention functions to model the semantic relationship between sentence pairs from different perspectives. 
%The two attentions are dot-product attention, subtract-based attention respectively. 
The input of the dual attention module is a triple of $P$, $Q$, $V$ $\in R^{d_{seq}\times d_{v}} $, where $d_{v}$ is the latent dimension, $d_{seq}$ is the length of the utterance. We use $p_i$, $q_i$ and $v_i$ to denote the $i$-th tokens of $P$, $Q$, and $V$ respectively. 
%Two independent attention mechanisms compute potential relationships between $Q$, $K$, and $V$ to measure their semantic interaction alignment. 
\vspace{-0.15cm}
\subsubsection{Dot Attention}
%Dot attention is part of the dual attention module, which can directly compute correlations using matrix operations, and the computed scores are correlation weights. It is also the most commonly used attention mechanism in semantic correlation modeling. And it follows the standard dot-product attention that the transformer operates by default. The input of the dot attention module consists of queries and keys of dimension $d_k$, and values of dimension $d_v$. We compute the dot products of the query with all keys, divide each by $\sqrt{d_k}$, and apply a softmax function to obtain the weights on the values. For the sake of simplicity, the formulations of BERT not be repeated here, please refer to \cite{devlin2018bert} for more details. We denote the output vector as:
%Dot attention is part of the dual attention module, which can directly compute correlations using matrix operations, and the computed scores are correlation weights. It is also the most commonly used attention mechanism in semantic correlation modeling. And it follows the standard dot-product attention that the transformer operates by default. The input of the dot attention module consists of queries and keys of dimension $d_k$, and values of dimension $d_v$. We compute the dot products of the query with all keys, divide each by $\sqrt{d_k}$, and apply a softmax function to obtain the weights on the values. For the sake of simplicity, the formulations of BERT not be repeated here, please refer to \cite{devlin2018bert} for more details. We denote the output vector as:
Dot attention is the most commonly used attention mechanism in semantic correlation modeling. And it follows the standard dot-product attention that the transformer operates by default. 
%The input of the dot attention module consists of queries and keys of dimension $d_k$, and values of dimension $d_v$. 
For the sake of simplicity, the formulations of it not be repeated here, please refer to \cite{devlin2018bert} for more details. We denote the output vector as:
\begin{footnotesize}
%\begin{subequations}
\begin{align}
& {s}_{j}^{t} = \mathbf{q}_{j}\odot \mathbf{v}_{t}, \quad a_{i}^{t} = \frac{exp(s_{i}^{t})}{\sum_{j=1}^{N}exp(s_{j}^{t})}, \quad \mathbf{q}_{t}^{d} =\sum_{i=1}^{N}a_{i}^{t}\mathbf{q}_{i}
\end{align}
%\end{subequations}
\end{footnotesize}
where $\mathbf{q}_{t}^{d} \in R^{1 \times d_{v}}$ is the output of the $t$-th position obtained after the dot attention calculation and $\odot$ is element-wise dot product.
 \vspace{-0.15cm}
\subsubsection{Subtract Attention}
%\subsubsection{Minus Attention}
%The third part of multi-view attention is the minus attention module that captures and aggregates the difference information between sentence pairs. The difference attention module adopts a subtraction-based cross-attention mechanism, which allows the model to pay attention to dissimilar parts between sentence pairs by element-wise subtraction as:
The second part of dual attention module is the subtract attention that captures and aggregates the difference information between sentence pairs. It allows the model to pay attention to dissimilar parts between sentence pairs by element-wise subtraction as:
\begin{footnotesize}
%\begin{subequations}
\begin{align}
& s_{j}^{t} = \tanh(\mathbf{W}_{m}(\mathbf{q}_{j}-\mathbf{v}_{t})), a_{i}^{t} = \frac{exp(s_{i}^{t})}{\sum_{j=1}^{N}exp(s_{j}^{t})}, \mathbf{q}_{t}^{s} =\sum_{i=1}^{N}a_{i}^{t}\mathbf{p}_{i}
\end{align}
%\end{subequations}
\end{footnotesize}
where $\mathbf{q}_{t}^{s} \in R^{1 \times d_{v}}$ is the output of the $t$-th position obtained after the minus attention calculation, and $\mathbf{W}_{m} \in R^{d_{seq} \times d_{v}}$ are trainable parameters.

\subsection{Composition Module}
The composition module is divided into two stages, one is internal aggregation and the other is external aggregation.
\vspace{-0.15cm}
\subsubsection{Internal Aggregation}
Internal aggregation is to integrate the representation obtained after attention with the original representation. For each position t, we concatenate ${v}_t$ with the representation ${q}_{t}^{c}$ obtained by attention, and then use gating to scale the overall information,c = (d,s). As shown below, This is an example of internal integration of dot attention:
\begin{footnotesize}
\begin{subequations}
    \begin{align}
        & \mathbf{x}_{t}^{d} = \left[q_{t}^{d},v_{t}\right],\quad g_{i} = \sigma\left(\textbf{W}_{g}\mathbf{x}_{t}^{d}\right)\\
        & \mathbf{x}_{t}^{d*} = g_{i}\odot \mathbf{x}_{t}^{d},\quad
        \mathbf{h}_{t}^{d} =\tanh(\mathbf{W}_{d}\mathbf{x}_{t}^{d*} + b_{d})
    \end{align}
\end{subequations}
\end{footnotesize}
For dot and subtract attention, we will also get $h_{t}^{d}$ and $h_{t}^{s}$, respectively. Where $\textbf{W}_{g} \in R^{1 \times 2d_{v}} $ , $\mathbf{W}_{d} \in R^{d_{v} \times 2d_{v}} $, $b_a$ are weights and bias of our model.

\vspace{-0.15cm}
\subsubsection{External Aggregation}
External aggregation is to fuse all the attention functions. We use a parameter $v_{i}$ as an input to adaptively fuse two different attention mechanisms.
\begin{footnotesize}
\begin{subequations}
\begin{align}
& s_{j} = v^{T}\tanh(\mathbf{W}_{1}h_{j}^{t} + \mathbf{W}_{2} \mathbf{v}_{j})(t = d,s)\\
& a_{i} =  \frac{exp(s_{i})}{\sum_{j=(d,s)}exp(s_{j})},\quad \mathbf{x}_{t} = \sum_{i=(d,s)}a_{i}\mathbf{h}_{t}^{i}
\end{align}
\end{subequations}
\end{footnotesize}

$X$=\{${x}_{1}$,...,${x}_{N}$\} is the final fused semantic feature. Finally, we feed $X$ into a multilayer perceptron (MLP) classifier for the probability pi of each label in the corresponding task. For all tasks, the objective function is to minimize the following cross entropy:
\begin{footnotesize}
\begin{equation}
\begin{aligned}
    \label{bert}
    \mathcal{L} &= \sum_{i=1}^{N}[y_{i}\log p_{i} + (1-y_{i})\log(1-p_{i})]
\end{aligned}
\end{equation}
\end{footnotesize}

where $y_{i}$ denotes a label, in paraphrase detection it is (0, 1) , in natural language inference it is the relation of two sentences of entailment, contradiction, and neutral.

\section{Experiments and Results}
\label{sec:exp}

\subsection{Datasets and Baselines}
\noindent \textbf{Datasets}  We conduct experiments on 10 sentence matching datasets to evaluate the effectiveness of our method. The GLUE~\cite{wang2018glue} benchmark is a widely-used dataset in thie field, which includes tasks such as sentence pair classification, similarity and paraphrase detection, and natural language inference\footnote{https://huggingface.co/datasets/glue}. We conduct experiments on 6 sentence pair datasets (MRPC, QQP, STS-B, MNLI, RTE, and QNLI) from GLUE. We also conduct experiments on 4 other popular datasets (SNLI, SICK, TwitterURL and Scitail). Furthermore, we tested the robustness of DPM using the Textflint\cite{gui2021textflint} tools.
%\footnote{https://huggingface.co/datasets/glue}
%\footnote{https://www.textflint.io}

\noindent \textbf{Baselines} To evaluate the effectiveness of our proposed DPM in SSM, we mainly introduce BERT \cite{devlin2018bert} and RoBERTa\cite{liu2019roberta}  for comparison. In addition, we also take competitive model transformer\cite{vaswani2017attention} without pre-training as baseline. In robustness experiments, we compare the performance of BERT on the robustness test datasets. For simplicity, the compared models are not described in detail here.

\subsection{Results and Analysis}
To evaluate the effectiveness of our method, we test the effectiveness of aggregating DPM in interaction-based and representation-based methods, respectively.

Firstly, we integrate DPM based on interaction-based methods. Table \ref{citation-guide-outsideGlue} shows the performance of DPM and competitive models on 10 datasets. It can be seen that the effect of non-pre-trained models is significantly worse than pre-trained models. This is mainly because the pre-trained model has more data from learning corpus and powerful information extraction ability.
When the backbone model is BERT-base or BERT-large, the average accuracy after integrating DPM is improved by 1.1\% and 0.7\%, respectively. The results show the effectiveness of our DPM framework on semantic matching tasks. Furthermore, our method outperforms RoBERTa-base by 1.5\% and RoBERTa-large by 0.6\%, respectively. which demonstrates that DPM can effectively capture the relationship between sentences from different aspects, so that more fine-grained and complex relationships can be exploited. Besides, the aggregated representation module can effectively fuse information from different attention modules.

Secondly, to verify the generalization performance of our method, we also test aggregated DPM among representation-based methods. The baseline model has the same settings as Sentence-BERT\cite{DBLP:journals/corr/abs-1908-10084}. 
%&It obtains two representations of two texts separately through the encoder, then concatenates the two representations, and finally uses a nonlinear function to distinguish the relationship.
%
The results are shown in Table \ref{citation-guide-outsideGlue2}. It can be seen that the representation-based method performs significantly worse than the interaction-based model.This is mainly because interaction-based methods can learn the alignment between phrases in a sentence and can better model sentence-pair relationships. And in the Scitail dataset, due to the small amount of training set data in Scitail, the variance of the model prediction results is large. However, DPM-Net still shows very competitive performance on the Scitail dataset. Furthermore, DPM-Net outperforms vanilla Bert and other competing models on almost all datasets. Those improvements demonstrate the benefit of dual-path modeling for mining semantics.

Overall, consistent conclusions can be drawn from these results. Compared with previous work, our method shows very competitive performance in judging semantic similarity, and the experimental results also confirm our method.
\begin{figure}[H]

\centering
\includegraphics[width=0.4\textwidth]{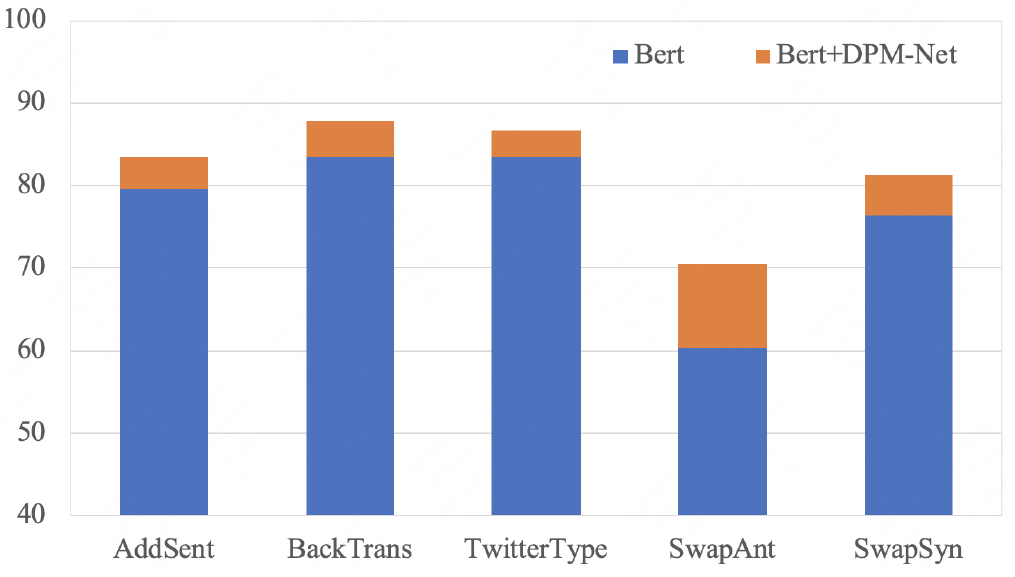}
\caption{\label{fig:robustess} The robustness experiment with DPM-Net  on SNLI dataset .
}
\vspace{-0.8cm}
\end{figure}
\subsection{Robustness Test Performance}
We conducted robustness tests on SNLI dataset. Table \ref{fig:robustess} lists the accuracy of DMP-Net and baseline model. We can observe that SwapAnt leads to a drop in maximum performance, and our model outperforms Bert nearly 10\% on SwapAnt, which indicates that DMP-Net can better handle semantic contradictions caused by antonyms.  And the model performance drops to 76.2\% on SwapSyn transformation, while DPM-net outperforms BERT by nearly 5\% because it requires the model to capture subtle entity differences for correct linguistic inference. In other transformations, DPM-Net still better than baseline, which reflects the advantages of dual path modeling in capturing subtle differences.
\begin{table}[H]
\vspace{-0.2cm}
\caption{\label{citation-guide-ablation}
Results of component ablation experiment.
}
\centering
\renewcommand\arraystretch{0.95}
\scalebox{0.8}{
\setlength{\tabcolsep}{4.5mm}{
\begin{tabular}{lcc|cc}
\toprule
%\hline
\multirow{2}*{Model} &\multicolumn{2}{c}{Quora} &\multicolumn{2}{c}{QNLI} \\ 
  \cmidrule(r){2-5}
  ~ & \text{Dev} & \text{Test} & \text{Dev} & \text{Test} \\
\midrule
%\hline
\textbf{DPM-Net} & \textbf{85.6} & \textbf{84.4}  & \textbf{88.3} & \textbf{86.0}\\
\text{w/o \ \ Dot-attention} & 84.5 & 83.2  & 87.1 & 84.9\\
\text{w/o \ \ Subtract-attention} & 85.1 & 83.5  & 87.3 & 85.2\\
\text{w/o \ \ Dual Attention } & 83.9 & 82.8  & 86.5 & 84.7\\
\text{w/o \ \ Internal Fusion } & 85.3 & 83.8 & 87.7 & 85.6 \\
\text{w/o \ \ External Fusion } & 85.4 & 83.9 & 87.9 & 85.7 \\
\bottomrule
%\hline
\end{tabular}}}
\vspace{-0.2cm}
\end{table}

\subsection{Ablation Study}
The experimental results are shown in the table \ref{citation-guide-ablation}.
First, remove dual Attention or remove the subcomponents in dual attention, the performance of the model on both datasets is significantly decreased. Which demonstrates the effectiveness of the internal components of the dual attention module.
Next, after removing internal fusion or external fusion, the performance of the model decreases by 0.5\% and 0.6\%, which proves that dynamic aggregation according to different weights can further improve the performance of the model.
Overall, due to the effective combination of each component, DPM-Net can adaptively fuse difference features into models and leverage its powerful contextual representation to better inference about semantics.

\vspace{-0.2cm}
\section{Conclusion}
\vspace{-0.1cm}
%In this paper, a two path modeling network is proposed to improve the ability of the model to perceive subtle differences in sentences.
%Experimental results on 10 public datasets and robustness dataset show that our method can achieve better performance than several strong baselines. 
In this paper, we propose a novel Dual Path Modeling Network (DPM-Net), which can efficiently aggregate the difference information in sentence pairs. DPM-Net enables the model to learn more fine-grained comparative information and enhances the sensitivity of models to subtle differences. Experimental results show that our method can achieve better performance than several strong baselines. Since DPM-Net is an end-to-end training component, it is expected to be applied to other large-scale pre-trained models in the future.

%In this paper, we propose a family of simple pooling front-ends (SimPFs) for improving the computation efficiency of audio neural networks. SimPFs utilize simple non-parametric pooling methods (e.g., max pooling) to eliminate the temporally redundant information in the input mel spectrogram. SimPFs achieve a substantial improvement in computation efficiency for off-the-shelf audio neural networks with negligible degraded or improved classification performance on four diverse audio classification datasets. In future work, we will study parametric pooling audio front-ends to adaptively reduce audio redundancy for efficient audio classification.
% and investigate the usefulness of our proposed method for Transformer-based audio neural networks \cite{gong2021ast}.  
\label{sec:conclusion}
\vspace{-0.2cm}
\section*{\normalsize ACKNOWLEDGEMENT}
\vspace{-0.1cm}
This work was supported by the National Key Research and Development Program of China (No. 2021YFB1714300) and National Natural Science Foundation of China (No.62006012) 

% \vfill\pagebreak
\bibliographystyle{IEEEtran}
\bibliography{strings,refs}

\begin{thebibliography}{10}
\providecommand{\url}[1]{#1}
\csname url@samestyle\endcsname
\providecommand{\newblock}{\relax}
\providecommand{\bibinfo}[2]{#2}
\providecommand{\BIBentrySTDinterwordspacing}{\spaceskip=0pt\relax}
\providecommand{\BIBentryALTinterwordstretchfactor}{4}
\providecommand{\BIBentryALTinterwordspacing}{\spaceskip=\fontdimen2\font plus
\BIBentryALTinterwordstretchfactor\fontdimen3\font minus
  \fontdimen4\font\relax}
\providecommand{\BIBforeignlanguage}[2]{{%
\expandafter\ifx\csname l@#1\endcsname\relax
\typeout{** WARNING: IEEEtran.bst: No hyphenation pattern has been}%
\typeout{** loaded for the language `#1'. Using the pattern for}%
\typeout{** the default language instead.}%
\else
\language=\csname l@#1\endcsname
\fi
#2}}
\providecommand{\BIBdecl}{\relax}
\BIBdecl

\bibitem{madnani-etal-2012-examining}
\BIBentryALTinterwordspacing
N.~Madnani, J.~Tetreault, and M.~Chodorow, ``Re-examining machine translation
  metrics for paraphrase identification,'' in \emph{Proceedings of the 2012
  Conference of the North {A}merican Chapter of the Association for
  Computational Linguistics: Human Language Technologies}.\hskip 1em plus 0.5em
  minus 0.4em\relax Montr{\'e}al, Canada: Association for Computational
  Linguistics, Jun. 2012, pp. 182--190. [Online]. Available:
  \url{https://aclanthology.org/N12-1019}
\BIBentrySTDinterwordspacing

\bibitem{bowman2015large}
S.~R. Bowman, G.~Angeli, C.~Potts, and C.~D. Manning, ``A large annotated
  corpus for learning natural language inference,'' \emph{arXiv preprint
  arXiv:1508.05326}, 2015.

\bibitem{wang2020multi}
S.~Wang, Y.~Lan, Y.~Tay, J.~Jiang, and J.~Liu, ``Multi-level head-wise match
  and aggregation in transformer for textual sequence matching,'' in
  \emph{Proceedings of the AAAI Conference on Artificial Intelligence},
  vol.~34, no.~05, 2020, pp. 9209--9216.

\bibitem{chen2016enhanced}
Q.~Chen, X.~Zhu, Z.~Ling, S.~Wei, H.~Jiang, and D.~Inkpen, ``Enhanced lstm for
  natural language inference,'' \emph{arXiv preprint arXiv:1609.06038}, 2016.

\bibitem{tay2017compare}
Y.~Tay, L.~A. Tuan, and S.~C. Hui, ``A compare-propagate architecture with
  alignment factorization for natural language inference,'' \emph{arXiv
  preprint arXiv:1801.00102}, vol.~78, p. 154, 2017.

\bibitem{devlin2018bert}
J.~Devlin, M.-W. Chang, K.~Lee, and K.~Toutanova, ``Bert: Pre-training of deep
  bidirectional transformers for language understanding,'' \emph{arXiv preprint
  arXiv:1810.04805}, 2018.

\bibitem{liu2019roberta}
Y.~Liu, M.~Ott, N.~Goyal, J.~Du, M.~Joshi, D.~Chen, O.~Levy, M.~Lewis,
  L.~Zettlemoyer, and V.~Stoyanov, ``Roberta: A robustly optimized bert
  pretraining approach,'' \emph{arXiv preprint arXiv:1907.11692}, 2019.

\bibitem{martins2016softmax}
A.~Martins and R.~Astudillo, ``From softmax to sparsemax: A sparse model of
  attention and multi-label classification,'' in \emph{International conference
  on machine learning}.\hskip 1em plus 0.5em minus 0.4em\relax PMLR, 2016, pp.
  1614--1623.

\bibitem{qiu2015convolutional}
X.~Qiu and X.~Huang, ``Convolutional neural tensor network architecture for
  community-based question answering,'' in \emph{Twenty-Fourth international
  joint conference on artificial intelligence}, 2015.

\bibitem{conneau2017supervised}
A.~Conneau, D.~Kiela, H.~Schwenk, L.~Barrault, and A.~Bordes, ``Supervised
  learning of universal sentence representations from natural language
  inference data,'' \emph{arXiv preprint arXiv:1705.02364}, 2017.

\bibitem{liang2019asynchronous}
D.~Liang, F.~Zhang, Q.~Zhang, and X.-J. Huang, ``Asynchronous deep interaction
  network for natural language inference,'' in \emph{Proceedings of the 2019
  Conference on Empirical Methods in Natural Language Processing and the 9th
  International Joint Conference on Natural Language Processing
  (EMNLP-IJCNLP)}, 2019, pp. 2692--2700.

\bibitem{miller1995wordnet}
G.~A. Miller, ``Wordnet: a lexical database for english,'' \emph{Communications
  of the ACM}, vol.~38, no.~11, pp. 39--41, 1995.

\bibitem{bodenreider2004unified}
O.~Bodenreider, ``The unified medical language system (umls): integrating
  biomedical terminology,'' \emph{Nucleic acids research}, vol.~32, no.
  suppl\_1, pp. D267--D270, 2004.

\bibitem{liang2019adaptive}
D.~Liang, F.~Zhang, W.~Zhang, Q.~Zhang, J.~Fu, M.~Peng, T.~Gui, and X.~Huang,
  ``Adaptive multi-attention network incorporating answer information for
  duplicate question detection,'' in \emph{Proceedings of the 42nd
  International ACM SIGIR Conference on Research and Development in Information
  Retrieval}, 2019, pp. 95--104.

\bibitem{zhang2020semantics}
Z.~Zhang, Y.~Wu, H.~Zhao, Z.~Li, S.~Zhang, X.~Zhou, and X.~Zhou,
  ``Semantics-aware bert for language understanding,'' in \emph{Proceedings of
  the AAAI Conference on Artificial Intelligence}, vol.~34, no.~05, 2020, pp.
  9628--9635.

\bibitem{liu2023time}
Y.~Liu, D.~Liang, F.~Fang, S.~Wang, W.~Wu, and R.~Jiang, ``Time-aware multiway
  adaptive fusion network for temporal knowledge graph question answering,''
  \emph{arXiv preprint arXiv:2302.12529}, 2023.

\bibitem{xia2021using}
T.~Xia, Y.~Wang, Y.~Tian, and Y.~Chang, ``Using prior knowledge to guide
  bert’s attention in semantic textual matching tasks,'' in \emph{Proceedings
  of the Web Conference 2021}, 2021, pp. 2466--2475.

\bibitem{bai2021syntax}
J.~Bai, Y.~Wang, Y.~Chen, Y.~Yang, J.~Bai, J.~Yu, and Y.~Tong, ``Syntax-bert:
  Improving pre-trained transformers with syntax trees,'' \emph{arXiv preprint
  arXiv:2103.04350}, 2021.

\bibitem{wang2022dabert}
S.~Wang, D.~Liang, J.~Song, Y.~Li, and W.~Wu, ``Dabert: Dual attention enhanced
  bert for semantic matching,'' in \emph{Proceedings of the 29th International
  Conference on Computational Linguistics}, 2022, pp. 1645--1654.

\bibitem{song-etal-2022-improving-semantic}
\BIBentryALTinterwordspacing
J.~Song, D.~Liang, R.~Li, Y.~Li, S.~Wang, M.~Peng, W.~Wu, and Y.~Yu,
  ``Improving semantic matching through dependency-enhanced pre-trained model
  with adaptive fusion,'' in \emph{Findings of the Association for
  Computational Linguistics: EMNLP 2022}.\hskip 1em plus 0.5em minus
  0.4em\relax Abu Dhabi, United Arab Emirates: Association for Computational
  Linguistics, Dec. 2022, pp. 45--57. [Online]. Available:
  \url{https://aclanthology.org/2022.findings-emnlp.4}
\BIBentrySTDinterwordspacing

\bibitem{gui2021textflint}
T.~Gui, X.~Wang, Q.~Zhang, Q.~Liu, Y.~Zou, X.~Zhou, R.~Zheng, C.~Zhang, Q.~Wu,
  J.~Ye \emph{et~al.}, ``Textflint: Unified multilingual robustness evaluation
  toolkit for natural language processing,'' \emph{arXiv preprint
  arXiv:2103.11441}, 2021.

\bibitem{li2021searching}
Z.~Li, J.~Xu, J.~Zeng, L.~Li, X.~Zheng, Q.~Zhang, K.-W. Chang, and C.-J. Hsieh,
  ``Searching for an effective defender: Benchmarking defense against
  adversarial word substitution,'' \emph{arXiv preprint arXiv:2108.12777},
  2021.

\bibitem{liu2021explainaboard}
P.~Liu, J.~Fu, Y.~Xiao, W.~Yuan, S.~Chang, J.~Dai, Y.~Liu, Z.~Ye, Z.-Y. Dou,
  and G.~Neubig, ``Explainaboard: An explainable leaderboard for nlp,''
  \emph{arXiv preprint arXiv:2104.06387}, 2021.

\bibitem{vaswani2017attention}
A.~Vaswani, N.~Shazeer, N.~Parmar, J.~Uszkoreit, L.~Jones, A.~N. Gomez,
  {\L}.~Kaiser, and I.~Polosukhin, ``Attention is all you need,'' in
  \emph{Advances in neural information processing systems}, 2017, pp.
  5998--6008.

\bibitem{wang2018glue}
A.~Wang, A.~Singh, J.~Michael, F.~Hill, O.~Levy, and S.~R. Bowman, ``Glue: A
  multi-task benchmark and analysis platform for natural language
  understanding,'' \emph{arXiv preprint arXiv:1804.07461}, 2018.

\bibitem{DBLP:journals/corr/abs-1908-10084}
\BIBentryALTinterwordspacing
N.~Reimers and I.~Gurevych, ``Sentence-bert: Sentence embeddings using siamese
  bert-networks,'' \emph{CoRR}, vol. abs/1908.10084, 2019. [Online]. Available:
  \url{http://arxiv.org/abs/1908.10084}
\BIBentrySTDinterwordspacing

\end{thebibliography}

\end{document}